\documentclass[10pt,twocolumn,letterpaper]{article}

\usepackage{cvpr}              %

\usepackage{marvosym}
\usepackage{capt-of}
\usepackage[dvipsnames]{xcolor}
\usepackage{xspace}
\usepackage{enumitem}
\usepackage{amsmath,amssymb,amsbsy,amsfonts,dsfont,pifont,bm,bbm,mathrsfs,mathtools,nicefrac}
\usepackage{algorithm,algpseudocode,listings}
\usepackage{booktabs,multirow,adjustbox,diagbox,threeparttable,tabularray}
\usepackage{colortbl}
\usepackage{arydshln}
\usepackage{makecell}

\usepackage[most]{tcolorbox}
\usepackage[table]{xcolor}  %
\newtcolorbox{promptbox}{
  colback=gray!5,
  colframe=gray!40,
  boxrule=0.4pt,
  arc=2pt,
  left=4pt,
  right=4pt,
  top=4pt,
  bottom=4pt,
}

\newcommand{\tocite}[1]{{\color{red} [TO CITE]}}
\newcommand{\method}{Alchemist\xspace}

\newcommand{\supp}{\textit{Supplementary Material}\xspace}

\definecolor{cvprblue}{rgb}{0.21,0.49,0.74}
\usepackage[pagebackref,breaklinks,colorlinks,citecolor=cvprblue]{hyperref}
\usepackage{wrapfig}
\usepackage{arydshln}
\usepackage{booktabs} 
\usepackage{graphicx}
\usepackage[capitalize]{cleveref}  %
\usepackage{arydshln}
\crefname{section}{Sec.}{Secs.}
\Crefname{section}{Section}{Sections}
\crefname{table}{Tab.}{Tabs.}
\Crefname{table}{Table}{Tables}
\crefname{figure}{Fig.}{Figs.}
\Crefname{figure}{Figure}{Figures}
\crefname{equation}{Eq.}{Eqs.}
\Crefname{equation}{Equation}{Equations}
\def\paperID{1721}
\def\confName{CVPR}
\def\confYear{2026}

\title{\includegraphics[height=6mm]{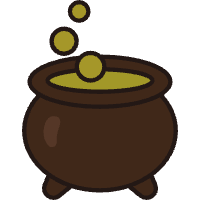}\ Alchemist: Unlocking Efficiency in Text-to-Image Model Training via Meta-Gradient Data Selection}
\author{%
Kaixin Ding\textsuperscript{$\rm 1^{\dagger}$}, 
Yang Zhou\textsuperscript{\rm 2}, 
Xi Chen\textsuperscript{\rm 1}, 
Miao Yang\textsuperscript{\rm 3},
Jiarong Ou\textsuperscript{\rm 3}, 
Rui Chen\textsuperscript{\rm 3}, 
Xin Tao\textsuperscript{\rm 3 \Letter},  
Hengshuang Zhao\textsuperscript{\rm 1 \Letter} \\
\textsuperscript{\rm 1}The University of Hong Kong,~\textsuperscript{\rm 2}South China University of Technology, ~\textsuperscript{\rm 3}Kling Team, Kuaishou Technology
}
\begin{document}

\twocolumn[{
\renewcommand\twocolumn[1][]{#1}
\maketitle
\begin{center}
    \includegraphics[width=1.0\linewidth]{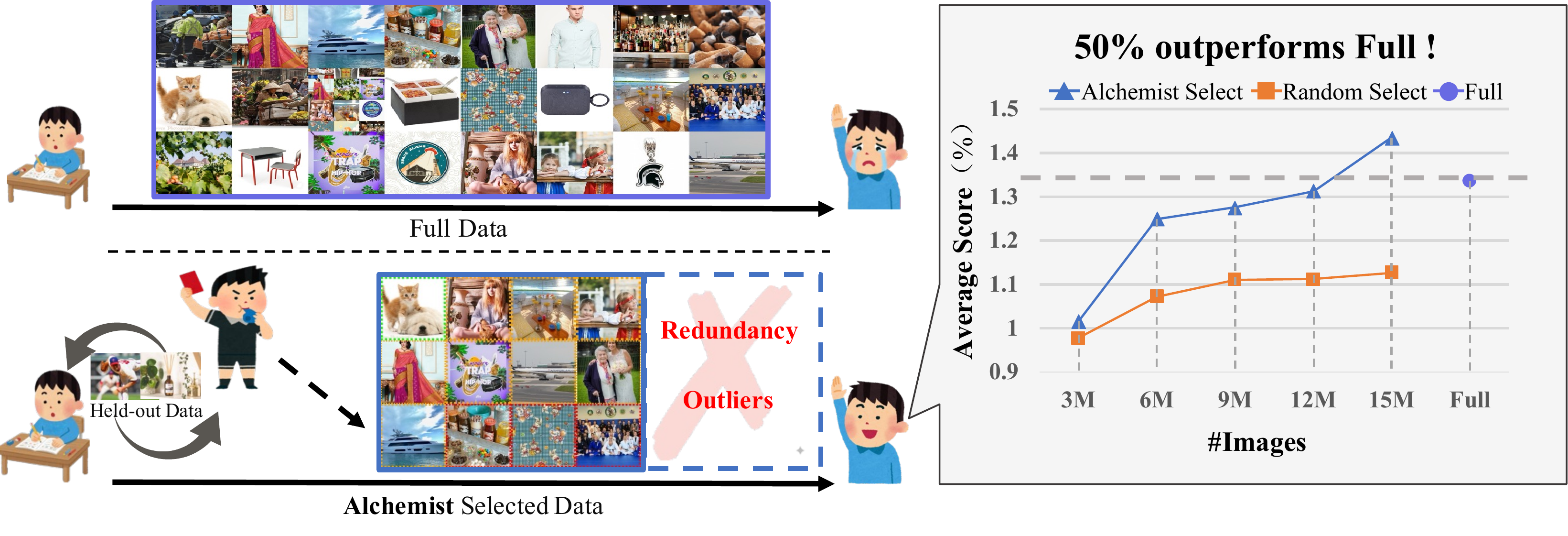}
    \vspace{-20pt}
    \captionsetup{type=figure}
    \caption{%
    \textbf{\method extracts informative data to improve data efficiency in Text-to-Image model training.}
We employ a rater model to score each sample and select the most informative subset for training. 
\method remains effective under arbitrary retention percentiles. Training on the \method-selected 50\% subset can even outperform training on the full dataset with the same number of training epochs.
Different rating are indicated by color-coded boxes.
Full refers to the 30M LAION~\cite{schuhmann2022laion} dataset.
The average score is computed as CLIP-Score divided by FID. Code is available on \href{https://kxding.github.io/project/Alchemist/}{\textbf{Project Page}}. 
}
    \label{fig:teaser}
\end{center}
}]

\let\thefootnote\relax
\footnotetext{\noindent\hspace{-1em}$\dagger$ Work done at Kling Team, Kuaishou Technology.}
\footnotetext{\noindent\hspace{-1em}\Letter~Corresponding Author}

\setlength{\abovedisplayskip}{3pt}
\setlength{\belowdisplayskip}{3pt}
\begin{abstract}
\vspace{-8pt}
Recent advances in Text-to-Image (T2I) generative models, such as Imagen, Stable Diffusion, and FLUX, have led to remarkable improvements in visual quality.
However, their performance is fundamentally limited by the quality of training data. 
Web-crawled and synthetic image datasets often contain low-quality or redundant samples which lead to degraded visual fidelity, unstable training, and inefficient computation. 
Hence, effective data selection is crucial for improving data efficiency.
Existing approaches rely on costly manual curation or heuristic scoring based on single-dimensional features in Text-to-Image data filtering. 
Although meta-learning based method has been explored in LLM, there is no adaptation for image modalities.
To this end, we propose \textbf{\method}, a meta-gradient-based framework to select a suitable subset from large-scale text-image data pairs. 
Our approach automatically learns to assess the influence of each sample by iteratively optimizing the model from a data-centric perspective. 
\method consists of two key stages: \textit{data rating} and \textit{data pruning}. 
We train a lightweight rater to estimate each sample’s influence based on gradient information, enhanced with multi-granularity perception. 
We then use Shift-Gsampling strategy to select informative subsets for efficient model training. 
\method is the first automatic, scalable, meta-gradient-based data selection framework for Text-to-Image model training. 
Experiments on both synthetic and web-crawled datasets demonstrate that \method consistently improves visual quality and downstream model performance. 
Training on an \method-selected 50\% of the data can outperform training on the full dataset.

\vspace{-13pt}
\end{abstract}

\section{Introduction}\label{sec:intro}

The remarkable progress of recent Text-to-Image (T2I) generative models, such as Imagen~\cite{saharia2022photorealistic}, Stable Diffusion~\cite{esser2024scaling}, and FLUX~\cite{flux2024}, has marked a major milestone in artificial intelligence, enabling the synthesis of highly realistic and semantically aligned images from textual descriptions. 
These models have demonstrated transformative potential across diverse domains, including art, design, and entertainment. 
However, the performance of such models is fundamentally bounded by the quality of their training data. 
While large-scale web-crawled~\cite{schuhmann2022laion} and synthetic image datasets~\cite{fang2025flux} provide abundant visual information, they inevitably contain low-quality samples—such as blurry images, plain-background advertisements, or other redundant content.
Training on such noisy or less-informative data can lead to degraded visual fidelity, training instability, and inefficient resource consumption, significantly hindering the scalability of T2I model development~\cite{abbas2023semdedup,slyman2024fairdedup}.
Therefore, effectively organizing and curating the training data is crucial for improving the performance of text-to-image models and thus enhancing data efficiency~\cite{albalak2024survey,xie2023data,gu2024data}.

Existing data selection approaches remain insufficient. 
Manual data curation is prohibitively expensive and lacks scalability.
Human-defined heuristic criteria~\cite{penedo2024fineweb,qiu2024wanjuan,wettig2024qurating} often rely on classifiers that evaluate data from one single dimension.
Although meta-learning-based methods~\cite{zhuang2025meta, calian2025datarater} have been explored in Large Language Model training, 
recent data selection approaches for Text-to-Image model training still rely predominantly on surface-form features~\cite{laion_aesthetic_predictor_2022}, \eg, CLIP-based~\cite{radford2021learning} linear estimators that predict aesthetic or clarity scores. 
Current methods fail to integrate complementary evaluation metrics. 
They also do not filter data from the perspective of maximizing downstream model performance, instead relying on human intuition.

To address these challenges, we propose \textbf{\method}, which is specially designed for large-scale text-image pairs data selection via meta-gradient.
Unlike heuristic scoring strategies, meta-learning simply requires users to specify \textit{what} they want at the end of training. 
The model then autonomously learns \textit{how} to value data samples through iterative meta-optimization, 
providing a scalable, data-driven approach to filter the training data stream.
Our framework consists of two key stages: \textbf{data rating} and \textbf{data pruning}. 
We first train a lightweight rater network to reweight every image item with predicted classification score based on gradient information extracted from the T2I proxy model to estimate each sample's influence. 
This design distinguishes images on their training dynamics~(\ie loss) with different ratings.
We also add multi-granularity perception to enable the rater to learn not only sample's own value but its contextual contribution within the group. 
Furthermore, we propose the Shift-GSample data pruning strategy which focuses on retaining the informative middle-to-late region---samples that are detailed yet learnable.
This strategy selects a subset of the rated dataset from a large corpus, thereby accelerating text-to-image model training and improving downstream task performance.
We apply \method to both synthetic image datasets and web-crawled datasets and also uses selected dataset to train on various model architectures.
\method consistently achieves higher performance under the same number of training epochs (see Fig.~\ref{fig:teaser}).
Moreover, we find that the selected subsets also align well with human intuitions, effectively filtering out images with plain backgrounds or highly chaotic content.
Our main contributions are summarized as follows:
\begin{itemize}[left=0pt]
\item \textbf{A general paradigm for text-to-image data selection.}  
We propose the first automatic, scalable, meta-gradient-based data selection framework for text-to-image model training. \method often selects a small subset of the data~(50\%) that outperforms training on the full dataset.

\item \textbf{Multi-granularity perception.}  
We design a rater that considers both individual sample features and their collective context, providing a more comprehensive and reliable assessment of data quality.

\item \textbf{Effective large-scale data selection.}  
We validate our \method selection method on four model families and three datasets of different domains. Our 15M selected subset enables the model to reach the same performance as random sampling but with 5× faster training time.
\end{itemize}
\section{Related Work}\label{sec:related}

\noindent \textbf{Coreset selection.} 
Coreset selection~\cite{phillips2016coresets,zheng2022coverage} aims to identify a small, representative subset of large-scale data that yields performance comparable to training on the full dataset. Many approaches rely on static heuristics. These include pruning data via predefined rules based on intrinsic properties (e.g., resolution, watermarks)~\cite{gururangan2020don, xia2020predicting, chen2023skill, zheng2024elfs, gu2024data}, or using pretrained models to score proxy attributes like aesthetic quality or text-image alignment~\cite{rombach2022high, yu2023quality, wang2025koala}. While effective for removing outliers, these strategies are limited because the proxy metrics are static and not directly optimized for the final downstream model's performance.
To address this gap, recent work has explored more dynamic, model-aware selection strategies. A promising direction leverages signals from the training process itself, such as model gradients, to identify high-impact data for pretraining~\cite{liu2024less,xia2024less}. This gradient-driven approach, shown to be effective in improving in-context learning, offers a more direct link between a sample's value and the learning objective. Building on this principle, our work adapts a meta-learning framework to the unique demands of T2I generation, learning a data-scoring function that is explicitly optimized for the generalization performance of the final generative model.

\noindent \textbf{Bilevel optimization for data selection.}
Model-based coreset selection methods~\cite{xia2024less, zhuang2025meta, calian2025datarater} that compute the exact gradient of validation performance are often formulated as a bilevel optimization problem. While recent works~\cite{calian2025datarater, nikdan2025efficient, dai2025data} adopt second-order derivatives to find precise solutions, these methods are computationally expensive and thus infeasible for large-scale modalities such as text-to-image (T2I) models.
To overcome this limitation, \method is instead closely related to penalty relaxation methods~\cite{clarke1990optimization, shen2023penalty, liu2022bome, kwon2023penalty, xiao2023generalized, lu2023slm, shen2025principled}, which efficiently approximate bilevel optimization solutions. Our work is most similar to \citet{shen2024seal}, but we introduce a continuous scoring mechanism that evaluates each sample based on the alignment between its individual gradient and the validation gradient. This design facilitates highly efficient adaptation to new downstream tasks.
\begin{figure*}[t!]
\centering 
\includegraphics[width=0.99\linewidth]{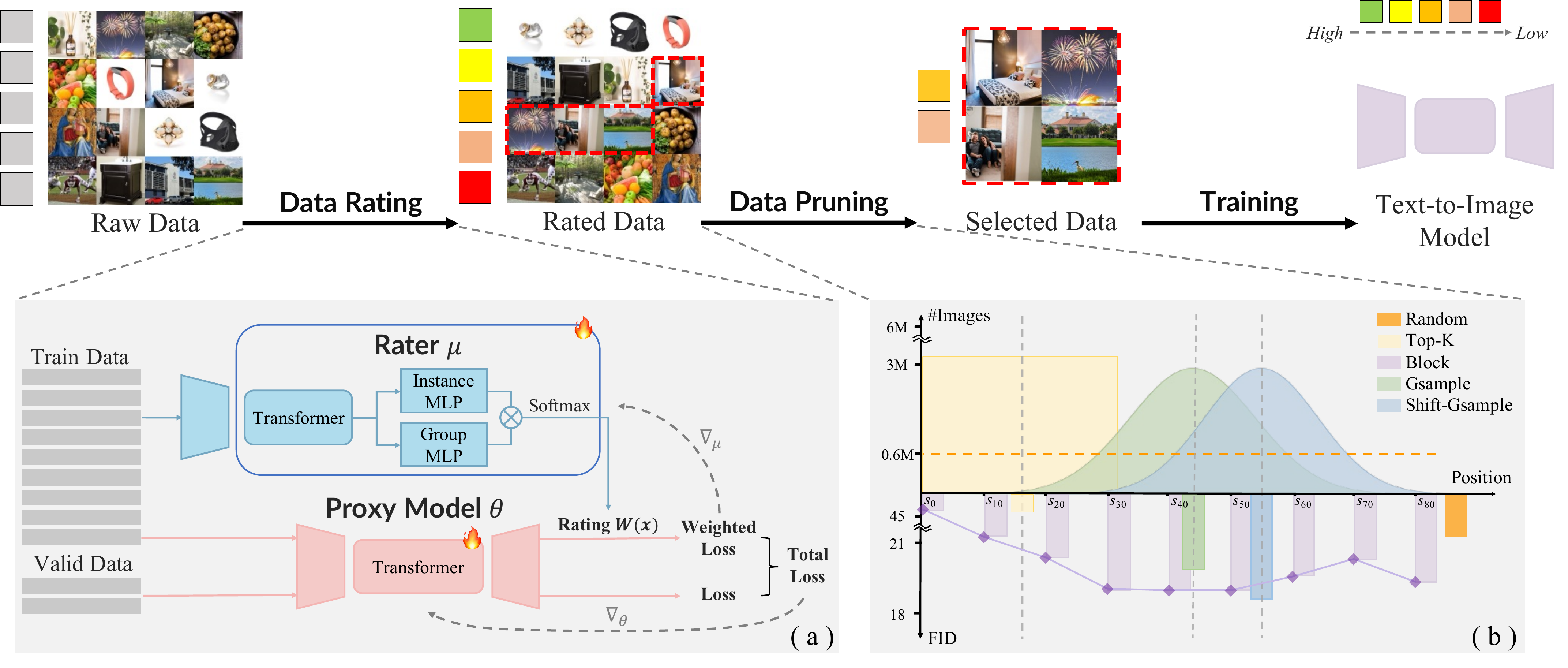} 
\vspace{-6pt}
\caption{\textbf{Overall pipeline of \method.} 
In the initial data rating stage~(a), the rater predicts a classification score for each image based on gradient extracted from a T2I proxy model. 
The rater and the proxy model are jointly optimized through weighted loss and total loss. 
In the data pruning stage~(b), we introduce the Shift-Gsample strategy to efficiently retain informative samples while filtering out redundant data and outliers. 
The resulting \method-selected dataset enables highly efficient training of downstream text-to-image models.
}
\label{fig:pipe_overal}
\vspace{-5pt}
\end{figure*}

\section{Method}
\label{sec:method}
Data efficiency lies in selecting the smallest or highest quality subset of the corpus that still yields strong results~\cite{albalak2024survey,xie2023data,gu2024data}.
Data selection aims to drop redundancy and outlier samples. 

Formally, our objective is to find a text-image pairs subset of full dataset \ie \(S \subset D\) such that the model \(\theta\) trained on \(S\) achieves better or comparable evaluation performance \(M\) than that trained on \(D\) with the same training epoch:
\begin{equation}
M\bigl(\theta(S)\bigr) \gtrsim M\bigl(\theta(D)\bigr).
\end{equation}
$\theta$ can be T2I models with different architectures.
We have proposed \method, as illustrated in Fig.~\ref{fig:pipe_overal}, which consists of two principal stages: data rating and data pruning.
Data rating based on gradient information extracted from a T2I proxy model to estimate each data's influence and data prunning chooses the suitable subset from the rated dataset to achieve data efficiency.

\subsection{Data Rating via Meta-Learning}

\noindent \textbf{Task formulation as bilevel optimization.}
The influence of each training sample can be quantified by its impact on a model’s performance on a held-out validation set~\cite{calian2025datarater,pruthi2020estimating}. 
Let $\theta \in \mathbb{R}^{d_{\theta}}$ denote the parameters of a proxy Text-to-Image (T2I) model, where $d_\theta$ is the number of learnable parameters.. 
Instead of selecting a discrete subset of data, we generalize this concept by learning a continuous weight $W_{x_i} \in [0,1]$ for each training instance $x_i \in \mathcal{D}_{\text{train}}$. 
The objective is to determine an optimal weighting scheme that minimizes the loss on a validation set $\mathcal{D}_{\text{val}}$.

To achieve this, we introduce a rating network~(rater) parameterized by $\mu \in \mathbb{R}^{d_{\mu}}$, where $d_{\mu}$ is the number of learnable parameters of the rater. 
Given a training sample $x_i$, the network outputs a continuous classification score $W_{x_i}(\mu)$. 
The goal is to learn the optimal rater parameters $\mu^*$ that minimize the validation loss of the proxy T2I model. 
This naturally leads to a bilevel optimization problem formulated as:
\begin{equation}
    \mu^* = \arg\min_{\mu} 
    \mathbb{E}_{x \sim \mathcal{D}_{\text{val}}}
    [\mathcal{L}(\theta^*(\mu); x)].
    \label{eq:bilevel_objective}
\end{equation}
The inner optimization seeks the optimal proxy model parameters:
\begin{equation}
    \theta^*(\mu) = 
    \arg\min_{\theta} 
    \sum_{x_i \in \mathcal{D}_{\text{train}}} 
    W_{x_i}(\mu)\,\mathcal{L}(\theta; x_i),
    \label{eq:inner_optimization}
\end{equation}
where $\theta^*(\mu)$ is implicitly a function of rater parameters $\mu$.

Directly solving the bilevel problem is computationally prohibitive, as it requires retraining the proxy T2I model to convergence for every update of the rater parameters.

\noindent \textbf{Meta-gradient optimization algorithm.}
To circumvent the challenge of solving the bilevel optimization problem, we employ an efficient gradient-based meta-learning strategy. This approach avoids fully solving the inner optimization by unrolling the training process for a few steps to approximate the meta-gradient for the rater's parameters $\mu$.

The optimization process, illustrated in Fig.~\ref{fig:pipe_overal} (a), proceeds as follows. First, we warm up the T2I proxy model to establish a stable initialization. 
During this phase, a reference proxy model with parameters $\hat{\theta}$ is updated using only the training data:
\begin{equation}
    \hat{\theta}_{k+1} = \hat{\theta}_k - \beta_k \nabla_{\theta} \mathcal{L}(\hat{\theta}_k; \mathcal{D}_{\text{train}}),
    \label{eq:theta_warmup}
\end{equation}
where $\beta_k$ is the learning rate at step $k$. After the warm-up period, the primary model parameters, $\theta$, are initialized with $\hat{\theta}$ and are subsequently updated using gradients from both the training and validation sets:
\begin{equation}
    \theta_{k+1} = \theta_k - \beta_k \left( g_{\text{val}}(\theta_k) + g_{\text{train}}(\theta_k, \mu_k) \right),
    \label{eq:theta_update}
\end{equation}
where $g_{\text{val}}(\theta_k) = \nabla_{\theta} \mathcal{L}(\theta_k; \mathcal{D}_{\text{val}})$ and $g_{\text{train}}(\theta_k, \mu_k) = \sum_{x_i \in \mathcal{D}_{\text{train}}} W_{x_i}(\mu_k) \nabla_{\theta} \mathcal{L}(\theta_k; x_i)$. In parallel, the reference parameters $\hat{\theta}_k$ continue to be updated using only the re-weighted training samples.
Concurrently, the rater's parameters $\mu$ are updated. Based on the theoretical framework of~\cite{shen2023penalty}, we employ an approximate stochastic gradient descent rule for the rater:
\begin{equation}
    \mu_{k+1} = \mu_k - \alpha_k 
    \left[
        \mathcal{L}(\theta_k; x_i) - \mathcal{L}(\hat{\theta}_k; x_i)
    \right]
    \nabla_{\mu} W_{x_i}(\mu_k),
    \label{eq:mu_update}
\end{equation}
where $\alpha_k$ is the meta learning rate and $x_i$ is a sample from $\mathcal{D}_{\text{train}}$. This update rule encourages the rater to assign higher weights to training samples that cause a larger reduction in the validation-informed loss relative to the training-only loss.

To ensure training stability, the sample weights $W_{x_i}$ are normalized within each batch. The rater outputs a raw score $\hat{W}_{x_i}$ for each sample, which is then passed through a softmax function:
\begin{equation}
    W_{x_i} =
    \frac{\exp(\hat{W}_{x_i})}
         {\sum_{j} \exp(\hat{W}_{x_j})},
    \label{eq:softmax_weight}
\end{equation}
where the summation is over all samples $x_j$ in the current batch.

In current SOTA T2I models, the parameter space is often significantly larger than the number of training samples ($d_{\theta} \gg |\mathcal{D}_{\text{train}}|$), so it is reasonable to assume the existence of model parameters $\theta^*$ that can achieve near-zero loss on the entire training set, i.e., $\mathcal{L}(\theta^*; x_i) \approx 0$ for all $x_i \in \mathcal{D}_{\text{train}}$.

This assumption implies that the inner optimization problem (Equation~\ref{eq:inner_optimization}), if solved to convergence, would yield a near-zero objective value regardless of the weights:
\begin{equation}
\begin{split}
    \min_{\theta} \sum_{x_i \in \mathcal{D}_{\text{train}}} W_{x_i}(\mu)\mathcal{L}(\theta; x_i) \\
    = \sum_{x_i \in \mathcal{D}_{\text{train}}} W_{x_i}(\mu)\mathcal{L}(\theta^*; x_i) \approx 0.
\end{split}
\end{equation}
In our meta-learning framework, the reference model $\hat{\theta}_k$ is trained exclusively on the training data and thus could serve as an approximation of these optimal parameters $\theta^*$. Consequently, we can posit that $\mathcal{L}(\hat{\theta}_k; x_i) \approx 0$ for a training sample $x_i$. Substituting this approximation into the rater update rule (Equation~\ref{eq:mu_update}) simplifies the gradient estimation significantly:
\begin{equation}
    \mu_{k+1} = \mu_k - \alpha_k \mathcal{L}(\theta_k; x_i) \nabla_{\mu} W_{x_i}(\mu_k).
    \label{eq:mu_update_simplified}
\end{equation}
This simplified rule provides a more direct and efficient update mechanism for the rater in over-parameterized settings.

\noindent \textbf{Multi-granularity Perception.}
\label{sec:batch}
Computing gradients over the entire training dataset $\mathcal{D}_{\text{train}}$ 
is computationally infeasible in practice. 
Therefore, we approximate the bilevel optimization using random minibatches. 
At each iteration $t$, a minibatch 
$B_t = \{x_1, \ldots, x_N\} \subset \mathcal{D}_{\text{train}}$ 
of size $N$ is sampled from the training set, 
and the gradient of the weighted loss is computed as:
\begin{equation}
    g_t = \nabla_{\theta} 
    \left( \frac{1}{N} \sum_{i=1}^{N} 
    W_{x_i}(\mu) \, 
    \mathcal{L}(\theta; x_i) \right),
    \label{eq:mini_grad}
\end{equation}
where $W_{x_i}(\mu)$ denotes the learned weight for instance $x_i$ 
output by the rater network parameterized by $\mu$.

The use of minibatches inevitably introduces stochastic bias, 
as some batches may contain samples of generally higher or lower quality than others. 
To mitigate this instability, we introduce a lightweight 
Group MLP module within the rater network, 
which captures the global characteristics of each minibatch.
Given the pooled instance features,
we first derive a batch-level representation by concatenating the mean and variance of all features.
This representation is passed through a two-layer MLP with sigmoid activation to produce a scalar batch weight $W_{\text{batch}}$.
Each instance feature is further processed by an Instance MLP and normalized via softmax to obtain instance weights $W_{\text{inst},i}$.
The final weight for each instance is computed as the product $W_i = W_{\text{inst},i} \cdot W_{\text{batch}}$.
This multi-level weighting enables the rater to capture both instance-level distinctiveness and batch-level coherence.
\begin{figure*}[t]
\centering 
\includegraphics[width=0.99\linewidth]{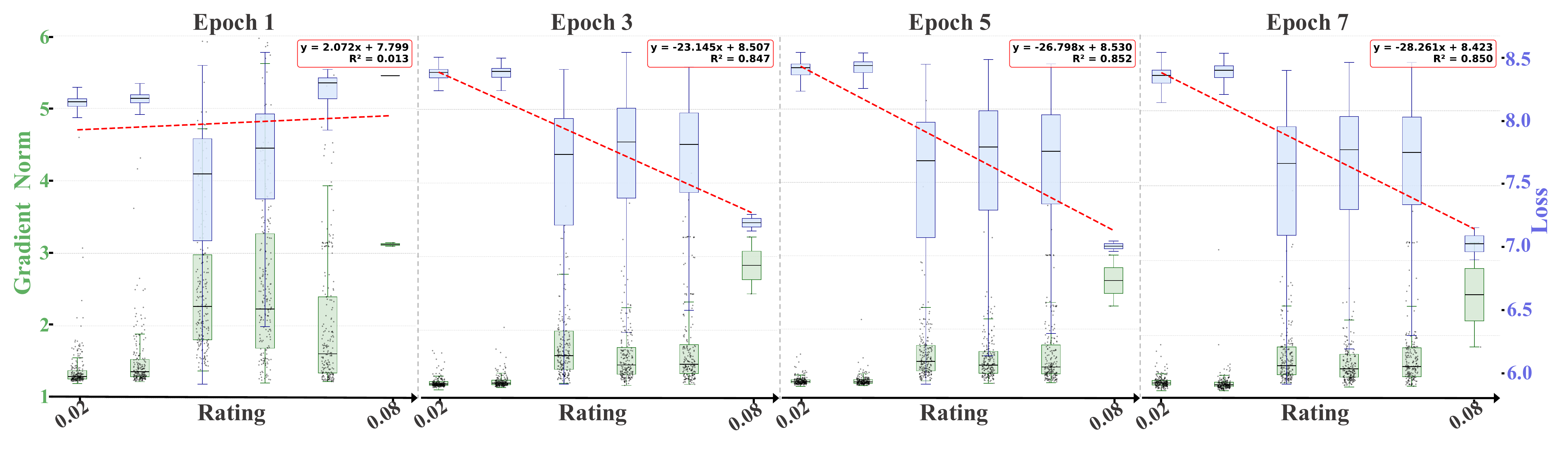} 
\vspace{-10pt}
\caption{
\textbf{Loss and gradient norm across different rating score ranges.}  
For each training sample, we record its instantaneous loss and gradient norm
at each training step during STAR-0.3B training.  
We track the evolution of loss and gradient norm over epochs.
}
\label{fig:loss-grad-rate}
\vspace{-8pt}
\end{figure*}

\begin{figure}[t]
\centering 
\includegraphics[width=0.99\linewidth]{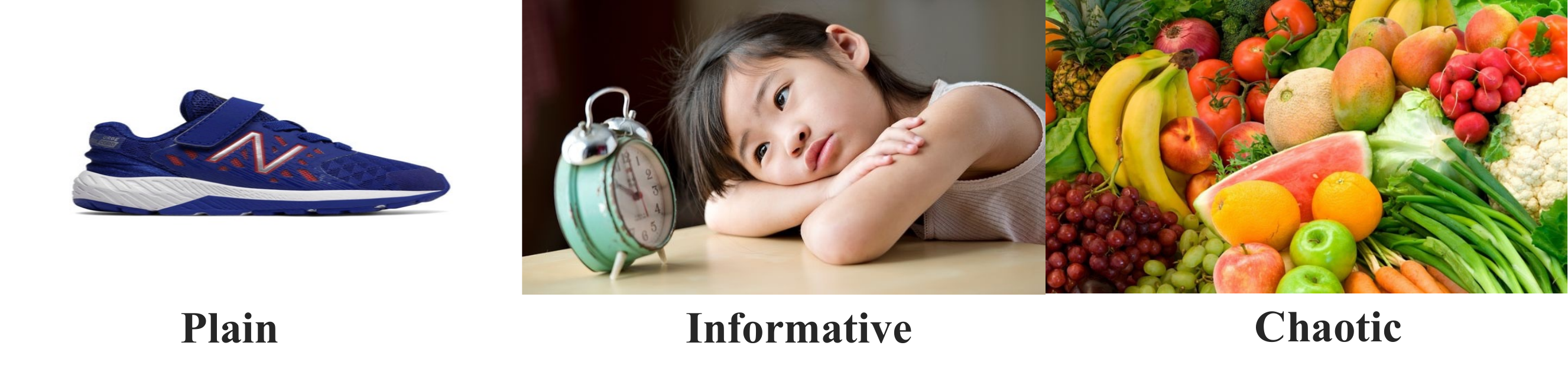} 
\vspace{-10pt}
\caption{
\textbf{Representative examples of data distribution across score regions.}  
The head region mainly contains \textit{plain} samples, the middle-to-late region contains \textit{informative} samples, and the tail region contains \textit{chaotic} samples.  
\method-selected data aligns with human intuition, filtering out most plain and chaotic samples.
}
\label{fig:case}
\vspace{-10pt}
\end{figure}
\subsection{Data Pruning Strategy}
With the rated data sorted from high to low, existing data selection methods typically adopt a Top-$K$ pruning scheme.  
However, for text-image data, we observe that simply keeping top-ranked samples does not always lead to better performance.
As rapid convergence may lead the model to overfit rather than achieve actual performance gains.

To analyze this, we randomly sample 10k data pairs across score ranges and track their training loss $\mathcal{L}_i(t)$ and gradient norm $g_i(t)=\|\nabla_{\theta}\mathcal{L}_i(t)\|_2$ during proxy model training.  
As shown in Fig.~\ref{fig:loss-grad-rate}, top-ranked samples maintain consistently low loss but exhibit minimal gradient change, indicating limited learning contribution.  
In contrast, samples within the mid-to-late score range show more active gradient dynamics and thus contribute more effectively to learning.
At the tail of the distribution, samples' gradients hardly descend.
Motivated by these findings, we argue that the most performance-contributing data lie in the mid-to-late portion of the distribution, while the highest-scoring region should be pruned out.

Based on these observations, we propose the pruning-based shift-Gaussian sampling~(\textbf{Shift-Gsample}) strategy that better balances data informativeness and diversity.
Specifically, after discarding the top $n\%$ of the sorted dataset, we perform Gaussian sampling with a shifted mean over the remaining samples:
\begin{equation}
    p(i) \propto 
    \exp\!\left(-\frac{(w_{x_i} - \mu)^2}{2\sigma^2}\right),
    \quad 
    w_{x_i} \in [n\%, 100\%],
\end{equation}
where $w_{x_i}$ denotes the normalized weight (percentile) of sample $x_i$ in the sorted dataset, $\mu$ represents the target mean percentile for sampling, and $\sigma$ controls the sampling spread.
We validate these insights experimentally in Fig.~\ref{fig:pipe_overal}~(b).
Five pruning methods are compared, each selecting an equal number of samples (20\% of the total number, \ie, 6M samples in our setting).
Random denotes uniform sampling independent of data ranking, while Top-$K$ selects the highest-scoring samples.
Block refers to a contiguous segment from the sorted list (\eg, $s_0$ starts at percentile $0\%$ ).
Gsample is the Gaussian Sample with the mean at the mid.
Our proposed shift-Gsample achieves the best performance improvement.
The trend of the \textcolor[HTML]{E1D5E7}{Block} method indicates that the most informative region lies in the mid-to-late range (around $s_{40}$--$s_{60}$),  
yet restricting sampling to this narrow segment reduces diversity---something shift-Gsample naturally balances.

We further provide insights into data distributions across different regions, as illustrated in Fig.~\ref{fig:case}.
Samples in the head region are typically easy and uninformative, while those in the tail are noisy or overly complex.
Our pruning strategy focuses on retaining the informative middle-to-late region—samples that are detailed yet learnable—striking a balance between learnability and diversity.
As a result, the selected subset aligns well with human intuition, effectively removing overly plain or chaotic images and leading to a more robust and generalizable training set for downstream T2I models.

\section{Experiments}
\label{sec:exp}
\subsection{Experimental Setup}
\begin{figure*}[t]
\centering 
\includegraphics[width=0.99\linewidth]{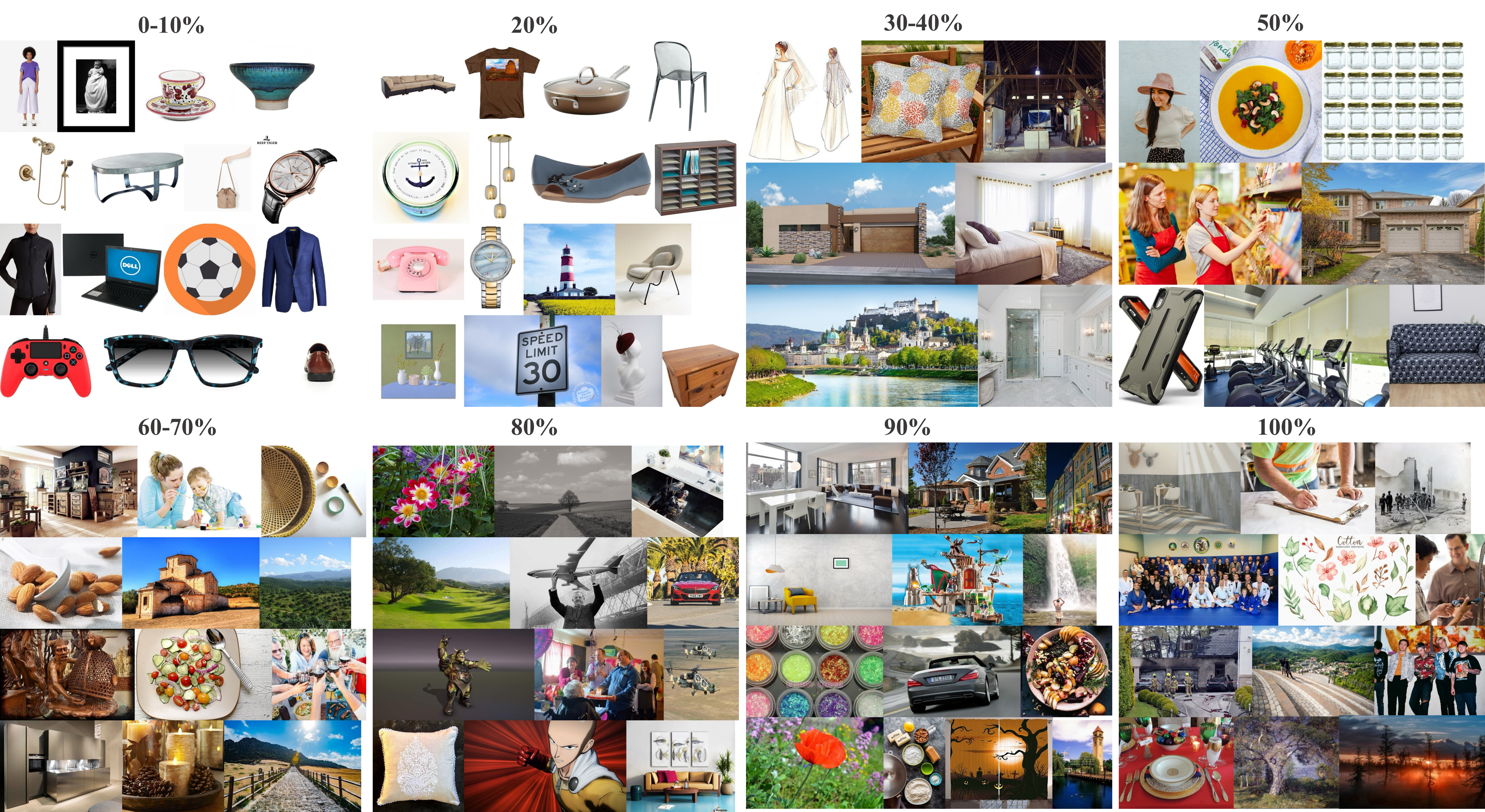} 
\caption{
\textbf{Image distribution of \method-selected LAION data subsets.}  
Samples are sorted from high to low scores.  
The early portion mainly contains images with white or plain backgrounds,  
the middle portion is more informative and content-rich,  
and the tail portion gradually becomes noisier, containing unclear content, multiple objects, or visually dense regions.
}
\label{fig:dataset_rated}
\vspace{-8pt}
\end{figure*}

\noindent \textbf{Training datasets.}
We use the following text–image datasets for data selection and model evaluation:  
(1) LAION~\cite{schuhmann2022laion}-30M, a large-scale web-crawled corpus with broad coverage but contains a considerable proportion of low-quality samples. Unless otherwise specified, LAION serves as the primary source for our experiments;
(2) Flux-reason~\cite{fang2025flux}-6M, a high-quality synthetic dataset generated by the FLUX model, specifically designed to capture reasoning-oriented image–text pairs; and  
(3) HPDv3~\cite{ma2025hpsv3}, a mixed real–synthetic dataset containing 1.17M annotated pairwise samples. It integrates both high-quality real photographs and images produced by state-of-the-art generative models, while retaining examples from older models and lower-quality real data to ensure diversity.  
We construct HPDv3-2M by pairing each text prompt with both human-preferred and non-preferred images, thereby introducing a mixture of samples with different tastes.

\noindent \textbf{Evaluation metrics.}
We evaluate the trained T2I models on MJHQ-30K~\cite{li2024playground} and report quantitative metrics for both image fidelity and text–image alignment through FID and CLIP-Score.
Additionally, we use GenEval~\cite{ghosh2023geneval} to assess generation performance in complex multi-object reasoning scenarios. 
Due to the relatively low intrinsic quality of the original full datasets, the overall absolute metrics are lower than those of the base models.
Further analysis is provided in \supp.

\noindent \textbf{Models for data selection and training.}
We evaluate \method on four base models: STAR-40M, STAR-0.3B, STAR-0.9B~\cite{ma2024star}, as well as FLUX-mini-3B~\cite{Tang2024FluxKits}.
All STAR models are trained from scratch, whereas FLUX-mini is fine-tuned with LoRA.
The rater model uses the same backbone as STAR-40M for image feature extraction.
The T2I proxy model uses STAR-0.3B.
STAR is a scale-wise text-to-image generation model designed for multi-resolution text-conditioned synthesis.
Flux-mini is distilled from Flux-dev~\cite{flux2024} and pruned by reducing network depth.
We use the corresponding optimal resolution.

\noindent \textbf{Implementation details.}
All experiments are implemented in PyTorch on NVIDIA A800 GPUs.  
The rater model is trained with a batch size of 64 across 2 GPUs, while the downstream text-to-image models are trained with the maximum feasible batch size across 8 GPUs.
For fairness, we keep the batch size consistent across all comparison methods with the same model.
For evaluating different data-selection strategies, we use STAR-0.3B. STAR models are trained from scratch for three epochs, and FLUX-mini is fine-tuned for one epoch unless otherwise stated.
Additional implementation details are provided in the supplementary material.

\subsection{Main Results}
We present comparisons with other approaches in Tab.~\ref{tab:scoremethod}, main results of \method across different models in Tab.~\ref{tab:modelsize} and adaptation to different datasets in Tab.~\ref{tab:datasize}.
Key findings are summarized below.

\noindent \textbf{50\% outperforms full.}
We compare our \method with several methods, including using the full dataset~(Full), random sampling~(Random), and heuristic selection metrics based on four common image quality metrics---Aesthetic, Clarity, Frequency, and Edge Density.  
Detailed pruning strategies for each heuristic metric are provided in \supp. We report the best-performing results for each to ensure fairness.
As shown in Tab.~\ref{tab:scoremethod}, the 50\% subset selected by our method achieves comparable performance of training on the full dataset while requiring only half the data.
Moreover, using merely 20\% of the data selected by \method yields performance comparable to training with 50\% randomly sampled data, demonstrating substantially improved data efficiency.
\begin{table*}[t]
\centering
\small
\setlength{\tabcolsep}{12pt}
\caption{
    Comparison of different image data selection methods evaluated on the MJHQ-30K and GenEval benchmarks.  
All models are trained from scratch for three epochs. Additional results with more training epochs are provided in Fig.~\ref{fig:epoch}.  
Ours-small refers to a subset selected by \method with a smaller number of images.  
Due to the relatively lower quality of the original full datasets, the overall evaluation metrics tend to be lower.
}
\label{tab:scoremethod}
\begin{tabular}{l c c c cc c c}
\toprule
\multirow{2}{*}{\textbf{Method}} & \multirow{2}{*}{\textbf{\#Params}} & \multirow{2}{*}{\textbf{\#Images}}  & \multirow{2}{*}{\makecell{\textbf{Training Time} \\ (hours)}} & \multicolumn{2}{c}{\textbf{MJHQ-30K\cite{li2024playground}}} & & \textbf{GenEval\cite{ghosh2023geneval}} \\
\cmidrule(lr){5-6} \cmidrule(lr){8-8}
& & & &FID$\downarrow$ & CLIP-Score$\uparrow$ & & Score$\uparrow$ \\
\midrule
Full & 0.3B & 30M & 65.34 & 17.48 & 0.2336 & & \textbf{0.2752} \\
\hdashline
Random & 0.3B & 15M & 34.60 & 19.70 & 0.2220 & & 0.2632 \\
Aesthetic\cite{laion_aesthetic_predictor_2022} & 0.3B & 15M & 34.60 & \underline{17.36} & \underline{0.2299} & & 0.2604 \\
Clarity\cite{laion_aesthetic_predictor_2022} & 0.3B & 15M & 34.60 & 17.85 & 0.2261 & & 0.2251 \\
Frequency\cite{castleman1979digital} & 0.3B & 15M & 34.60 & 18.77 & 0.2276 & & 0.2519 \\
Edge-density\cite{otsu1975threshold} & 0.3B & 15M & 34.60 & 20.13 & 0.2240 & & 0.2429 \\
\hdashline
Ours-small & 0.3B & 6M & 13.08 & 18.22 & 0.2277 & & 0.2367 \\
Ours & 0.3B & 15M & 34.60 & \textbf{16.20} & \textbf{0.2325} & & \underline{0.2645} \\
\bottomrule
\end{tabular}
\vspace{-10pt}
\end{table*}

\noindent \textbf{Effectiveness across different models}
Tab.~\ref{tab:modelsize} presents comparison results against baseline random selection on the MJHQ-30K evaluation benchmark.
Using the rater to evaluate the dataset once, the selected data can be transferred to train any target model, whether training STAR models from scratch or fine-tuning FLUX models with LoRA adapters.
\method is also effective across different retention ratios, as shown in Fig.~\ref{fig:teaser}.
In all cases, \method consistently outperforms random selection, demonstrating that our data selection approach identifies informative samples and improves data efficiency for text-to-image models.
\begin{table}[t]
\centering
\small
\caption{
The values in \textcolor{ForestGreen}{green} superscript indicate the performance improvement compared to the random sampling in different model scales, model types and data scales. From-scratch training on STAR and LoRA finetuning on FLUX-mini.
}
\label{tab:modelsize}
\setlength{\tabcolsep}{9.5pt}
\begin{tabular}{cccc}
\toprule
\textbf{\#Params} & \textbf{\#Images} & \textbf{FID} ↓ & \textbf{CLIP-Score} ↑ \\
\midrule
\multicolumn{4}{l}{\textit{(a) Experiments on STAR with different model scale}} \\[3pt]
40M  & 6M & 26.74$^{\textcolor{ForestGreen}{\tiny -1.92}}$ & 0.2059$^{\textcolor{ForestGreen}{\tiny +0.0050}}$ \\

0.3B & 6M & 18.22$^{\textcolor{ForestGreen}{\tiny -2.47}}$ & 0.2277$^{\textcolor{ForestGreen}{\tiny +0.0058}}$ \\

0.9B & 6M & 16.71$^{\textcolor{ForestGreen}{\tiny -1.43}}$ & 0.2356$^{\textcolor{ForestGreen}{\tiny +0.0027}}$ \\
\midrule
\multicolumn{4}{l}{\textit{(b) Experiments on STAR with different data scale}} \\[3pt]
0.3B & 3M  & 21.54$^{\textcolor{ForestGreen}{\tiny -0.61}}$ & 0.2193$^{\textcolor{ForestGreen}{\tiny +0.0024}}$ \\

0.3B & 6M  & 18.22$^{\textcolor{ForestGreen}{\tiny -2.47}}$ & 0.2277$^{\textcolor{ForestGreen}{\tiny +0.0058}}$ \\

0.3B & 15M & 16.20$^{\textcolor{ForestGreen}{\tiny -3.50}}$ & 0.2325$^{\textcolor{ForestGreen}{\tiny +0.0105}}$ \\
\midrule
\multicolumn{4}{l}{\textit{(c) Experiments on FLUX-mini}} \\[3pt]
3B & 6M & 19.98$^{\textcolor{ForestGreen}{\tiny -0.43}}$ & 0.2366$^{\textcolor{ForestGreen}{\tiny +0.0007}}$ \\
\bottomrule
\end{tabular}
\vspace{-8pt}
\end{table}

\noindent \textbf{Scalability to larger and different model families.}
We demonstrate that using a small proxy model can effectively boost performance for both larger models and different model families.
As shown in Tab.~\ref{tab:modelsize}, we apply \method on LAION dataset and then train STAR-T2I-0.3B, STAR-T2I-0.9B (\ie, a larger model), and finetune FLUX-mini-3B (\ie, a different model family) for evaluation.
Across all settings, \method consistently outperforms random selection, indicating that the selected data transfers well across model scales.
Importantly, the training cost of the small rater model is negligible compared with training larger generative models, enabling data selection to be amortized across multiple downstream tasks.

\noindent \textbf{Adaptability across different data domains.}
\method also generalizes effectively across diverse data sources. We conduct experiments on HPDv3~\cite{ma2025hpsv3} and Flux-reason~\cite{fang2025flux}, comprising 2M and 6M image-text pairs respectively.
HPDv3 contains both real photographs and synthetic images, with human-preferred and non-preferred image-text pairs, while Flux-reason consists of entirely synthetic, reasoning-oriented data.
As shown in Tab.~\ref{tab:datasize}, our method outperforms the baseline at both 20\% and 50\% data retention percentage.
Combined with results on web-crawled dataset LAION from previous experiments, \method demonstrates strong adaptability to heterogeneous data domains.
\begin{table}[t]
\centering
\caption{
Comparison of multiple data ratention percentage on different domain datasets, HPDv3-2M and Flux-reason-6M.
}
\label{tab:datasize}
\small
\setlength{\tabcolsep}{2.5pt}
\begin{tabular}{lcccc}
\toprule
\multirow{2}{*}{\textbf{Data percentage}} & \multicolumn{2}{c}{\textbf{HPDv3-2M~\cite{ma2025hpsv3}}} & \multicolumn{2}{c}{\textbf{Flux-reason-6M~\cite{fang2025flux}}} \\
\cmidrule(lr){2-3} \cmidrule(lr){4-5}
 & FID ↓ & CLIP-Score ↑ & FID ↓ & CLIP-Score ↑ \\
\midrule
Random (20\%) & 35.55 & 0.1977 & 23.66 & 0.2135 \\[2.5pt]
Ours (20\%)   & \textbf{32.27} & \textbf{0.2039} & \textbf{22.78} & \textbf{0.2142} \\[2.5pt]
\hdashline 
\noalign{\vskip 2.5pt}
Random (50\%) & 20.21 & 0.2201 & 19.35 & 0.2260 \\[2.5pt]
Ours (50\%)   & \textbf{18.15} & \textbf{0.2265} & \textbf{18.59} & \textbf{0.2265} \\
\bottomrule
\end{tabular}
\vspace{-8pt}
\end{table}

\noindent \textbf{Efficiency in text-to-image model training.}
As shown in Fig.~\ref{fig:epoch}, STAR-0.3B achieves 2.33× and 5× faster training time using \method-selected data while reaching performance comparable to random-selected at 20\% and 50\% data retention percentage, respectively.
\subsection{Ablation Study}

\noindent \textbf{Multi-granularity perception.}
Due to memory constraints, we perform gradient updates within minibatches rather than on the entire dataset.
Differences between batches can introduce bias. To mitigate this, we incorporate a group MLP in the rater to capture global information across the dataset.
As shown in Tab.~\ref{tab:ablation_strategies}, considering this group perspective improves performance compared to only evaluating each data sample individually.

\noindent \textbf{Pruning strategies.} 
We evaluate three different pruning strategies in Tab.~\ref{tab:ablation_strategies}.
The Shift-Gsample strategy achieves the best performance, indicating that samples from the middle-to-late portions of the rated dataset yield the largest gains. These portions correspond to the samples exhibiting the fastest gradient changes, as shown in Fig.\ref{fig:loss-grad-rate}.
The Shift-Gsample pruning strategy, guided by the data rating, can be adapted across different domains to filter out redundant or less informative samples.

\begin{table}[t]
\centering
\caption{
Ablation study at 6M text-image pairs. 
}
\vspace{-8pt}
\label{tab:ablation_strategies}
\small
\setlength{\tabcolsep}{5pt}
\begin{tabular}{lccc}
\toprule
\textbf{Method} & \textbf{\#Images} & \textbf{FID} ↓ & \textbf{CLIP-Score} ↑ \\
\midrule
TopK      & 6M & 48.20 & 0.1945 \\
Gsample   & 6M & 19.22 & 0.2248 \\
Ours (Shift-Gsample)    & 6M & \underline{18.37} & \underline{0.2272} \\
\hspace{1em}+Group-MLP     & 6M & \textbf{18.22} & \textbf{0.2277} \\
\bottomrule
\end{tabular}
\vspace{-10pt}
\end{table}

\noindent \textbf{Training trends. }
\begin{figure}[t]
\centering 
\includegraphics[width=0.99\linewidth]{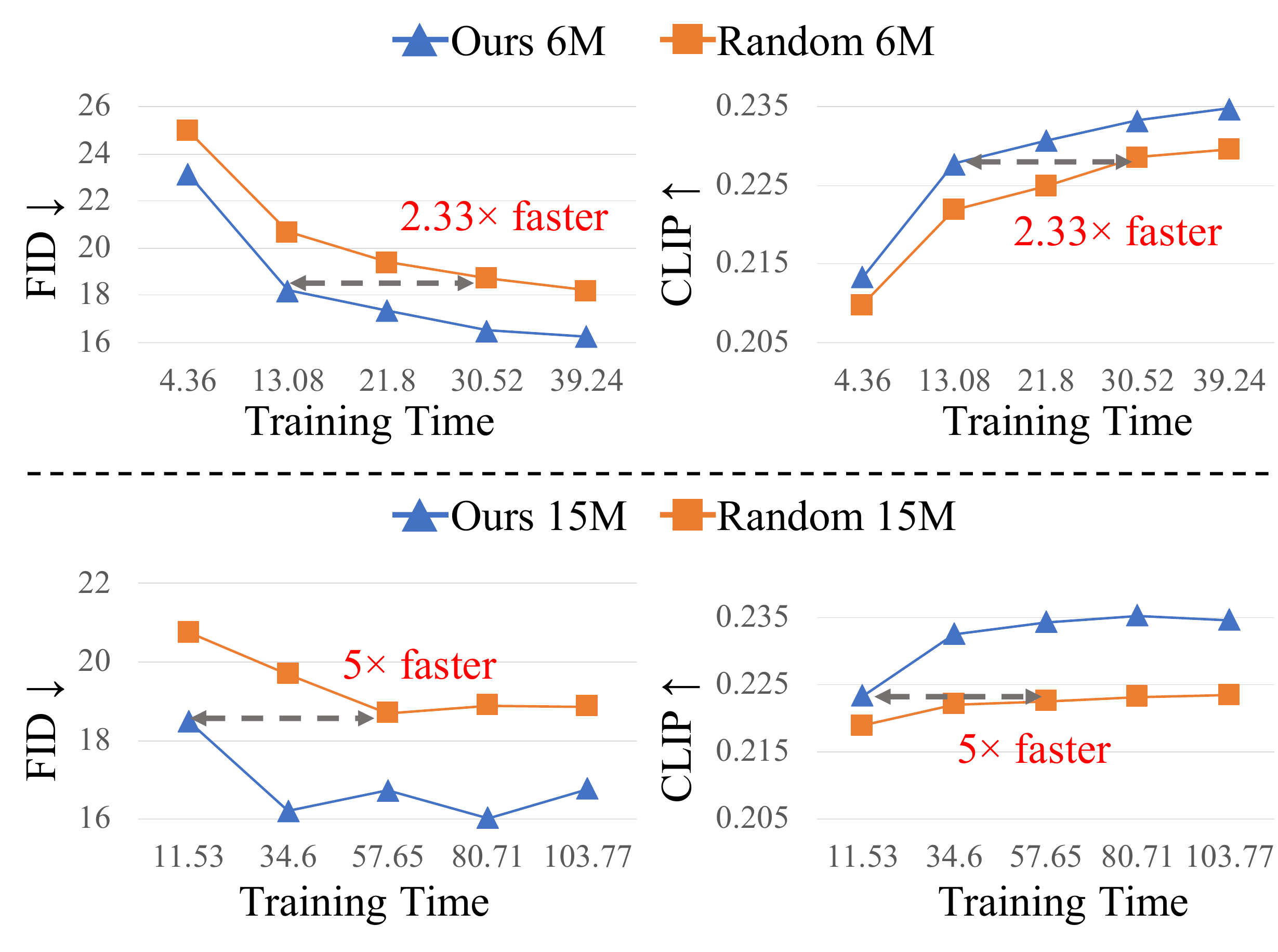} 
\vspace{-8pt}
\caption{
Performance of models trained on 6M and 15M data vs training time. Evaluations are conducted on MJHQ-30K benchmark.
}
\label{fig:epoch}
\vspace{-12pt}
\end{figure}
As shown in Fig.~\ref{fig:epoch}, we track training time and performance for models trained on 6M and 15M data, respectively.
\method consistently accelerates improvements throughout training.
In contrast, random sampling exhibits early-stage fluctuations and slower gains.
\method demonstrates steady and stable improvement, highlighting the effectiveness and robustness of our data selection approach.
As models approach convergence~(approximately 9-10 epochs), performance naturally fluctuates before stabilizing.
\section{Analysis}
\noindent \textbf{What does the \method-sorted dataset look like?}
We visualize the \method-sorted datasets in Fig.~\ref{fig:dataset_rated}, showing subsets corresponding to the 0\%-100\% percentiles of the ranked data. 
We observe that samples in the top-ranked subset (0-20\%) are typically \textit{clean, simple, and often with plain backgrounds}. While these samples lead to lower training loss, they lack semantic richness. 
In contrast, the middle-and-late ranked subsets (e.g., 40-80\%) contain higher-quality, informative samples with clear subjects and well-defined actions. 
Finally, the lowest-ranked subset (e.g., 90-100\%) predominantly consists of noisy, chaotic, cluttered scenes with multiple fine-grained or inconsistent subjects.

\noindent \textbf{What are the differences in image modality?}
Previous studies in LLMs have already explored meta-learning-based data selection~\cite{xia2024less}. But they only limited to one stage and simply rated data with lower loss and achieve the top part, assuming that lower loss leads to improved data efficiency. 
While clean text often indicates high information with no redundant tokens, it works in LLM. However this assumption does not directly hold for images.
In the image domain, a single image exhibits substantial redundancy~(up to 70\%~\cite{kersten1987predictability}), \eg large regions with plain backgrounds.
Consequently, highly rated images with lower loss tend to be visually clean but contain limited information, contributing little to model learning.
Although such samples may accelerate convergence, we observe that the resulting models are often overfitted and produce lower-quality outputs.
Focusing solely on minimizing the inner T2I proxy model loss can therefore introduce bias.
To this end, we propose \method. 
Our approach uses ratings to estimate each sample’s influence on convergence speed and applies pruning to balance convergence speed and final performance.
We also selectively retain primarily informative samples while including a small portion of easy or noisy examples to maintain data diversity.

\noindent \textbf{Why autoregressive proxy T2I model?}
While our T2I proxy model adopts an autoregressive (AR) structure, the particular modeling paradigm is not crucial, as long as it provides sufficient discrimination between data samples. 
We adopt an AR proxy primarily for its single-step generation process, which enables efficient gradient propagation and stable meta-optimization. 
This choice avoids the problems faced in diffusion models, where some intermediate timesteps results can correlate with good visual quality but still suffer from high noise and large loss, thus misleading the optimization signal.
\section{Conclusion}\label{sec:conclusion}
We present \method, the first automatic, scalable, and gradient-based data selection method specifically designed for real-world Text-to-Image model training.  
Extensive experiments demonstrate that \method consistently improves both training efficiency and model performance.  
Our method establishes a new paradigm for meta-learning based data curation in Text-to-Image development and opens up opportunities for larger text–image datasets in the future.
{\small
\bibliographystyle{conference}
\bibliography{conference.bib}
}
\newpage

\end{document}